\documentclass[letterpaper, 10 pt, conference]{ieeeconf}  % Comment this line out if you need a4paper

\IEEEoverridecommandlockouts                              % This command is only needed if 

\usepackage{balance}
\usepackage{cite}
\usepackage{graphicx}
\usepackage{url}
\usepackage{gensymb}
\usepackage{mathtools}
\usepackage{caption}
\usepackage{float}
\usepackage{booktabs} % To thicken table lines
\usepackage{multirow}
\title{\LARGE \bf
Grasp Type Estimation for Myoelectric Prostheses using \\Point Cloud Feature Learning
}

\author{\normalsize Ghazal Ghazaei$^{1,2}$, Federico Tombari$^{2}$, Nassir Navab$^{2}$ and Kianoush Nazarpour$^{1,3}$
	\thanks{This work is supported by UK Engineering and Physical Sciences Research Council (EP/R004242/1).}
	\thanks{$^{1}$ School of Engineering, Newcastle University, Newcastle-upon-Tyne NE1 7RU, UK.}
	\thanks{$^{2}$ Technical University of Munich, Munich, Germany.}
	\thanks{$^{3}$ Institute of Neuroscience, Newcastle University, Newcastle-upon-Tyne NE2 4HH, UK.}
	\thanks{Emails: \{g.ghazaei1\}@newcastle.ac.uk}}

\begin{document}

\maketitle

\thispagestyle{empty} %forces first page to not have a header

%\ieeefootline{Workshop on Human-aiding Robotics \\ International Conference on Intelligent Robots and Systems 2018}%creates footline

%%%%%%%%%%%%%%%%%%%%%%%%%%%%%%%%%%%%%%%%%%%%%%%%%%%%%%%%%%%%%%%%%%%%%%%%%%%%%%%%
\begin{abstract}
Prosthetic hands can help people with limb difference to return to their life routines. Commercial prostheses, however have several limitations in providing an acceptable dexterity. We approach these limitations by augmenting the prosthetic hands with an off-the-shelf depth sensor to enable the prosthesis to see the object's depth, record a single view (2.5-D) snapshot, and estimate an appropriate grasp type; using a deep network architecture based on 3D point clouds called PointNet. The human can act as the supervisor throughout the procedure by accepting or refusing the suggested grasp type. We achieved the grasp classification accuracy of up to 88\%. Contrary to the case of the RGB data, the depth data provides all the necessary object shape information, which is required for grasp recognition. The PointNet not only enables using 3-D data in practice, but it also prevents excessive computations. Augmentation of the prosthetic hands with such a semi-autonomous system can lead to better differentiation of grasp types, less burden on user, and better performance. 
\end{abstract}

\section{Introduction}
Losing a hand can cause inevitable limitations to an individual's life. Prosthetic hands can provide such amputees with the opportunity of returning to their normal activities. However, control of these prosthetic hands is still unnatural and limited to a few degrees of freedom. Therefore, research is ongoing to further improve the functionality of prosthetic hands~\cite{Nazarpour2014,Saunders2011,krasoulis2017improved,atzori2016deep,dovsen2010cognitive,ninu2014closed}.

There are several research works which employ 2-D and 3-D visual data to boost the performance of prosthetic hands, demonstrating the benefit of using vision as an additional modality to the electromyogram (EMG) signals~\cite{ninu2014closed,dovsen2010cognitive,Markovic2014,ghazaei2015exploratory,ghazaei2017deep,Giordaniello2017}. In \cite{dovsen2010cognitive}, 10 consecutive 2D RGB snapshots of objects together with ultrasound distance information are used as an input to a rule-based reasoning algorithm to estimate among four different grasp types. Later in~\cite{markovic2015sensor}, fusion of different sensory data including myoelectric recordings, computer vision, inertial measurements and embedded prosthesis sensors (position and force) led to semi-autonomous and proportional control of a prosthetic hand in multiple DOFs. RGB-D imaging was used to estimate the shape, size and orientation of objects. Another work that benefited from 3-D sensors was~\cite{ninu2014closed}, which proposed a combination of stereo-vision and augmented reality (AR) for better user interface and control of the hand. In~\cite{ghazaei2015exploratory,ghazaei2017deep}, an RGB image is fed to a two-layer convolutional neural network (CNN)~\cite{LeCun1995} to choose the best grasp among four different types. The algorithm can effectively classify objects based on their appropriate grasp category without any additional sensor or measurement. 
%All these works demonstrated considerable improvement in the performance of commercial hand prostheses. Yet, they are not benefiting from the most recent advancements of computer vision.

Most works indicate that the use of additional modalities such as depth can be beneficial in grasp estimation with vision. Despite the high grasp recognition performance in~\cite{ghazaei2015exploratory,ghazaei2017deep}, the system was sensitive to change in distance and view point, which can be overcome by using a depth sensor. Additionally, background removal is a challenging task in 2D images, while depth data can ease this procedure significantly and provide better outcomes. With the rapid development of 3-D computer vision and depth sensors, it is now much easier than before to use depth data and process it. There are a variety of sensors, which can provide depth for objects as close as 10 cm. They are adequately small (from 6-9 cm) and can be effectively integrated into the prosthesis. Intel\textregistered{} RealSense\textsuperscript{TM} D400 series~\footnote{Intel RealSense Depth Camera D400-Series:\\ \url{http://ufldl.stanford.edu/tutorial/supervised/ConvolutionalNeuralNetwork/}. Accessed: 2018-01-08.}, REAL3\textsuperscript{TM} image sensor family by Infineon Technologies AG~\footnote{3-D Image Sensor REAL3.\\ \url{https://www.infineon.com/cms/en/product/sensor/3d-image-sensor-real3/}. Accessed: 2018-01-09.} and Pico depth sensors by Pmdtechnologies AG\footnote{Worldclass 3-D depth sensing with Pico family. \url{http://pmdtec.com/picofamily/}. Accessed: 2018-01-09.} are among the 3-D sensors with the potential to be integrated into a prosthesis.

Although these innovations facilitate the use of depth sensors, 3-D data processing can be computationally expensive. A solution could be the PointNet~\cite{qi2016pointnet} approach, that relies on a comparatively shallow network. PointNet uses point cloud data directly and has shown great performance on several tasks including 3-D shape classification, which is of our interest~\cite{qi2016pointnet}. Another benefit is that the RGB data is no more required, which eliminates the use of unnecessary data and accelerates the performance. By employing the recent developments in computer vision, this paper tries to improve the grasping performance of artificial hands and presents an efficient semi-autonomous grasp estimation approach for a single view (2.5-D) point cloud (set of points), which can easily be implemented on an available artificial hand by a single depth sensor. That is, a depth sensor augmented on a prosthetic hand can capture a single view RGB-D image by a trigger command recorded from the amputee user. The image is processed, converted  to point cloud and fed into the PointNet, which results in an automatic grasp act in the prosthesis by classifying the object based on its appropriate grip pattern. As geometric information is substantially important for grasp gesture of an object, we extract the normal vectors of point clouds and input that to the PointNet as extra information. Results indicate estimation improvement of $\sim 10\%$ when normal data is added.

\section{Methods}

\subsection{Dataset}
\subsubsection{Washington RGB-D dataset}

\begin{figure}[!t]
 \begin{center}
  \includegraphics[scale=0.9]{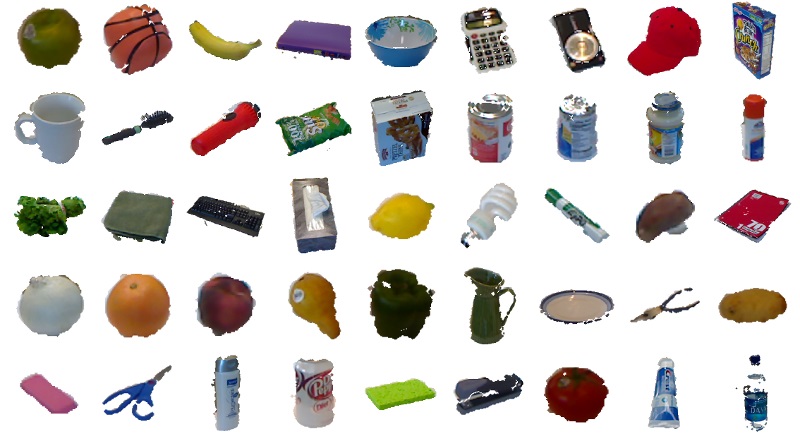}
  \setlength{\belowcaptionskip}{-25pt}
  \caption{Examples of objects from the Washington RGB-D dataset.}
  \label{figure_1}
  \end{center}
\end{figure}

There are numerous datasets, which exploit depth data. However, due to the specific aim of grasping in this paper, we focused on the ones that include graspable objects. Among those, a large RGB-D dataset collected at the University of Washington~\cite{lai2011large,WashingtonDataset} includes sufficient data and presents mostly graspable objects, for example those shown in Figure~\ref{figure_1}. There are RGB and depth images of 300 common everyday objects from multiple view angles (total: 250,000 RGB-D images) collected with a Microsoft Kinect\footnote{Microsoft Kinect \url{https://www.xbox.com/en-us/kinect}. Accessed: 2018-01-09}. We selected 3321 2.5-D point clouds, which are distributed in 48 categories and sampled from almost every $12$ views per object in each object category. In some categories, some objects may not be used to avoid unnecessary repetition.

We used the processed point clouds provided in~\cite{WashingtonDataset} as the background is already removed. We manually labeled these objects based on their appropriate grasp type following the process adopted in~\cite{ghazaei2015exploratory,ghazaei2017deep} to four different grasp groups: \textit{tripod}, \textit{pinch}, \textit{palmar wrist neutral} and \textit{palmar wrist pronated}.\par

\subsubsection{BigBIRD ((Big) Berkeley Instance Recognition Dataset)}

\begin{figure}[H]
 \begin{center}
  \includegraphics[scale=.8]{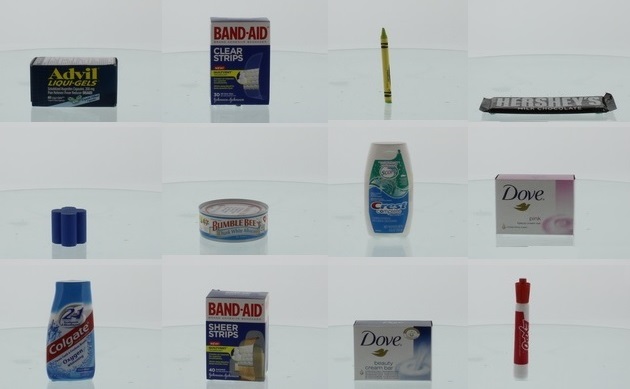}
  \setlength{\belowcaptionskip}{-15pt}
  \caption{The selected objects from the BigBIRD dataset.}
  \label{figure_2}
  \end{center}
\end{figure}

Since the Washington RGB-D dataset includes more objects in \textit{palmar wrist neutral} and \textit{palmar wrist pronated} than  \textit{tripod} and \textit{pinch} grasp categories, we added some data from the BigBIRD dataset~\cite{bigbird} to compensate and provide sufficient diversity in every grasp class. It includes 100 objects and 600 RGBD images for each. Here, we picked 12 categories, including 656 2.5-D point clouds sampled from processed files (background is removed). The RGB and depth data are collected by Canon and Primesense Carmine cameras. Figure~\ref{figure_2} depicts the objects selected from the BIGBIRD.

Combining these two datasets, Table~\ref{table_1} indicates the number of objects per grasp class in each group. It is worth noting that sample selection per available views rate in each object category in the BigBIRD is 5 times more than the Washington RGB-D dataset due to the presence of many similar objects in the latter. In this way, the overall number of point clouds per grasp is nearly uniformly distributed as shown in the last row of Table~\ref{table_1}.
\\

\begin{table}[H]
\vspace{-1em}
\caption{A summary of data distribution within grasp groups and datasets. Palmar WN and Palmar WP stand for \textit{palmar wrist neutral} and \textit{palmar wrist pronated} grasps respectively.}
\begin{center}
\begin{tabular}{lr@{\hskip 0.5em} r@{\hskip 0.5em} r@{\hskip 0.5em} r@{\hskip 0.5em} r@{\hskip 0.5em} r} 
\toprule
& & \multicolumn{4}{c}{Grasp Pattern}   \\ \cline{3-6}
 Dataset & Type  & Pinch & \begin{tabular}{@{}c@{}}Palmar \\ WN\end{tabular}  & Tripod & \begin{tabular}{@{}c@{}}Palmar \\ WP\end{tabular}  \\  \midrule
\multirow{ 2}{*}{Washington} & objects  & 62 & 80 & 15 & 82  \\ 
& point clouds & 740 & 956 & 644 & 981 \\ \midrule
\multirow{ 2}{*}{BIGBIRD} & objects & 4 & - & 8 & -  \\ 
& point clouds & 183 & - & 473 & - \\ \midrule
\multirow{ 2}{*}{Combined} & objects & 66 & 80 & 23 & 82 \\ 
& point clouds & 923  & 956 & 1117 & 981  \\ 
\bottomrule
\end{tabular}
\end{center}
\label{table_1}
\vspace{-1em}
\end{table}

\subsection{Data preparation}

Although the point clouds were already processed and the background was removed properly, more processing was required as the PointNet requires the point clouds to be zero-mean and normalized into an unit sphere. 

As an additional data relevant to object geometry, the surface normals for each point cloud were also estimated. Although there are several normal estimation methods available, one of the simplest approaches is to approximate the normal to a point on the surface by estimation of the normal of a plane tangent to the surface, which becomes a least-square plane fitting estimation problem. Consequently, the surface normal estimation problem is reduced to an analysis of the eigenvectors and eigenvalues of a covariance matrix created from the nearest neighbors of the query point. That is, for each point $\boldsymbol{p}_i$, the covariance matrix $\mathcal{C}$ can be calculated according to equation~\ref{eq_1}.
\begin{equation}\label{eq_1}
\mathcal{C} = \frac{1}{k}\sum_{i=1}^{k}{\cdot (\boldsymbol{p}_i-\overline{\boldsymbol{p}})\cdot(\boldsymbol{p}_i-\overline{\boldsymbol{p}})^{T}}, ~\mathcal{C} \cdot \vec{{\mathsf v}_j} = \lambda_j \cdot \vec{{\mathsf v}_j},~ j \in \{0, 1, 2\}
\end{equation}
where $k$ indicates the number of point neighbors considered in the neighborhood of $\boldsymbol{p}_i$ (here $k=100$ provided us with desirable results), $\overline{\boldsymbol{p}}$ illustrates the 3-D centroid of the nearest neighbors, $\lambda_j$ represents the $j$-th eigenvalue of the covariance matrix, and $\vec{{\mathsf v}_j}$ the $j$-th eigenvector~\cite{RusuDoctoralDissertation}.

Finally, we uniformly sampled 2048 points for each point cloud. It is worth noting that we did not use the RGB data as it does not include any shape relevant information and consequently barely any grasping relevant data. Thus, each point cloud is composed of six coordinates $(x,y,z,n_x,n_y,n_z)$, where $n_i$ represents the normal vector for $x,y,z$. For training, point clouds are augmented by random rotation along the up-axis and jittering the position of each point by a zero-mean Gaussian noise (standard deviation, 0.01).

\begin{figure}[t]
	\begin{center}
		\includegraphics[scale=1.05]{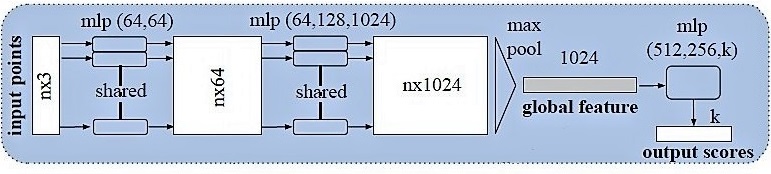}
		\caption{The original PointNet unified architecture for classification. Point clouds are directly taken as input and the output is a grasp class. When surface normals are also fed to the network, the input size increases to $n\times6$.}
		\label{figure_3}
	\end{center}
\end{figure}

\subsection{PointNet}
Deep learning methods have shown great success in various classification tasks~\cite{Krizhevsky2012,van2014analysis}. Although point clouds are simple and unified type of geometric data structure and easy to learn from, they are not directly fed to a deep network architecture due to their irregular format. PointNet however can simply use point clouds as the input representation~\cite{qi2016pointnet} thanks to its unique design (Figure~\ref{figure_3}).\par
Since a point cloud is a set of unordered 3-D points, PointNet requires certain symmetrization in the feed-forward computation and further invariances to rigid motions may also be needed. The main feature of PointNet is the presence of a single symmetric function called \textit{max pooling} that aggregates the information from each point leading to invariance to input permutations. As shown in Figure~\ref{figure_3}, the network selects informative points of the point cloud during training in the first MLP (multi-layer perceptron) layers. These learned optimal values are accumulated into the global descriptor by the final fully connected layers. For our specific task of grasp estimation, this global descriptor should include particular distinctive features that represent each grasp category.\par

Batch normalization~\cite{ioffe2015batch} and ReLU (rectified linear unit) are used for all the layers. We used the learning rate of 0.001 to train the network on an Nvidia Geforce GTX 960M GPU.

\section{Results and discussion}

\begin{figure}[H]
 \begin{center}
  \includegraphics[scale=0.35]{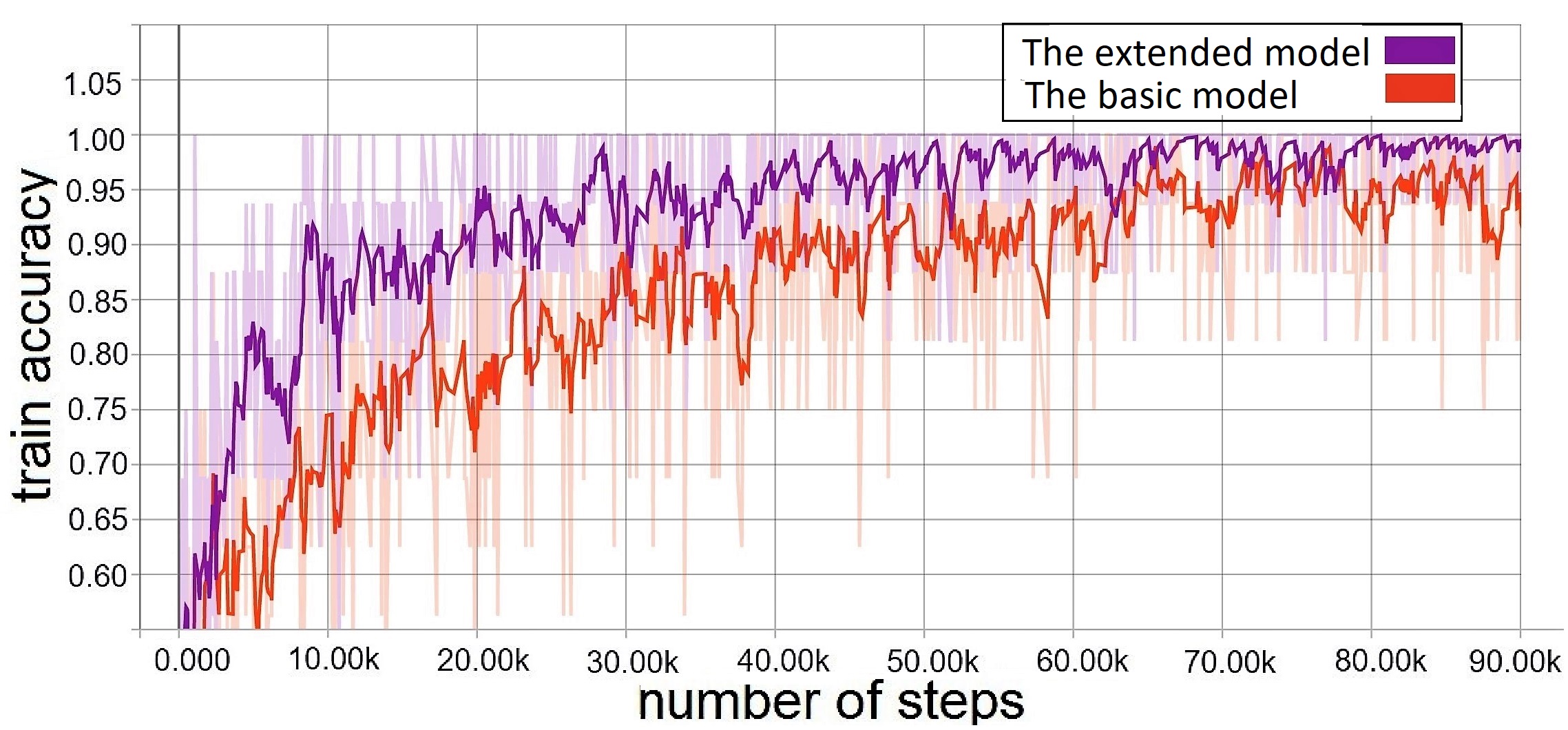}
  \caption{The accuracy curve for training of the a) basic and b) extended models. The exact same parameters are used for the two networks. }
  \label{figure_4}
  \end{center}
\end{figure}

We had a total of 3797 point clouds of which we used $80\%$ for training, $10\%$ for validation and the remaining $10\%$ for testing. We trained the PointNet in two ways: 1) basic model including $(x,y,z)$ data only and 2) Extended model including surface normals, $(x,y,z,n_x,n_y,n_z)$. Results are depicted in Table~\ref{table_2}. Figure~\ref{figure_4} illustrates the training curves for both models in the second fold of cross-validation. The extended model converges in fewer steps to a higher accuracy while taking longer training time.\par

\begin{table}[H]
\begin{center}
\caption{The PointNet performance in grasp estimation. Five-fold cross-validation results in terms of average accuracy and standard deviation are reported.}
\begin{tabular}{ lr@{\hskip 0.5em} r@{\hskip 0.5em} r } 
\toprule
 Grasp\textbackslash Model &  Basic model &  Extended model     \\  \midrule
 Pinch &                  $0.707\pm0.08$  &  $0.799\pm0.064$   \\
 Palmar wrist neutral &   $0.966\pm0.026$  &  $0.978\pm0.015$      \\
 Tripod &                 $0.72\pm0.089$  &  $0.822\pm0.039$    \\
 Palmar wrist pronated &  $0.795\pm0.042$ &  $0.826\pm 0.039$   \\ \midrule
 \textbf{Overall} & $\mathbf{0.793\pm0.021}$ & $\mathbf{0.854\pm0.025}$ \\
\bottomrule
\end{tabular}
\label{table_2}
\vspace{-1em}
\end{center}
\end{table}

According to table~\ref{table_2}, the results indicate about $79\%$ average accuracy for the basic model and $85\%$ average accuracy for the extended model. The procedure of processing an image and predicting relevant grasp for it takes about $0.03$ seconds. It can be seen that using surface normals as additional coordinates is beneficial to the grasp estimation task (performance improvement up to $\sim10\%$ in one of the cross-validation folds). It seems to be a plausible claim as surface normals can provide more data relevant to the object shape and grasping gesture. 

It can be observed that since the \textit{palmar wrist neutral} grasp type includes the most distinctive types of objects compared to other grasp groups (objects that their  length along $y$-axis is larger than their length along $x$-axis), the objects suitable for this grasp type are recognized with the highest accuracy. Moreover, the \textit{pinch} grasp consisting of the least amount of data represents the lowest recognition accuracy. These results also fit with previous results reported in~\cite{ghazaei2015exploratory,ghazaei2017deep}.

\begin{figure}[t]
 \begin{center}
  \includegraphics[scale=0.45]{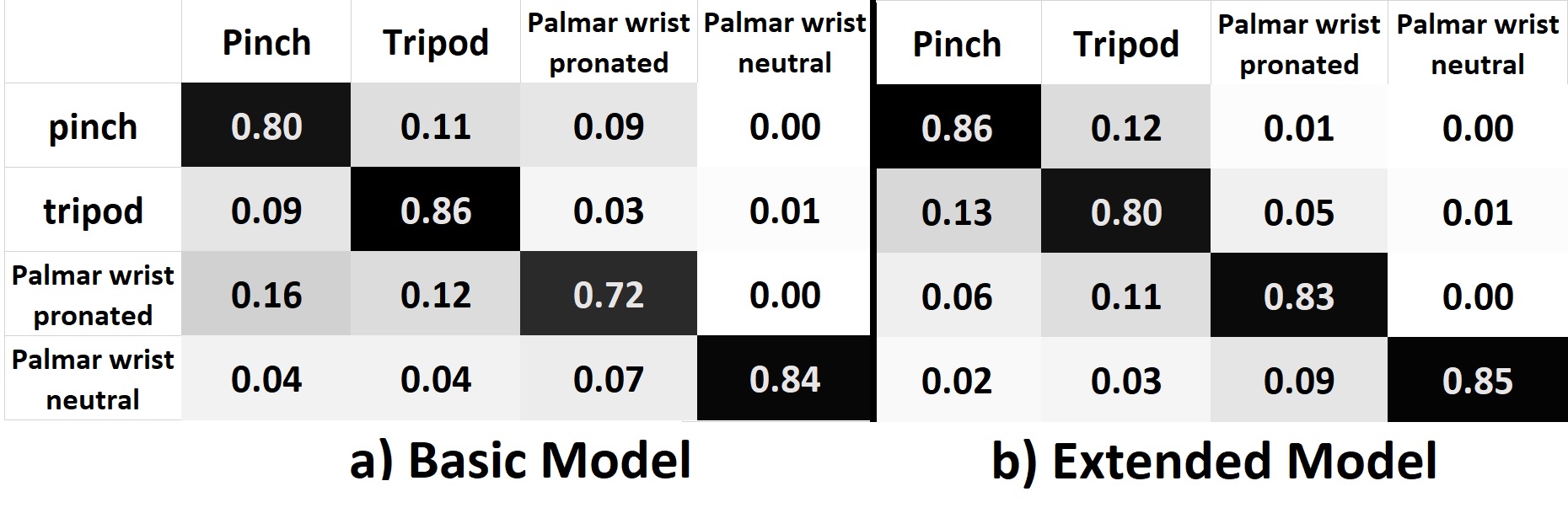}
  \caption{Illustration of confusion matrices for a)the basic and b)the extended models. The unacceptable errors (such as \textit{pinch} grasp mistaken by \textit{palmar wrist pronated} grasp) are more frequent in case of the basic model.}
  \label{figure_5}
  \end{center}
\end{figure}

Figure~\ref{figure_5} indicates the confusion matrices of the second validation fold for both basic and extended models. As results already presented, the extended model indicates a better distribution around the diagonal.

\begin{figure}[!htp]
 \begin{center}
  \includegraphics[scale=0.45]{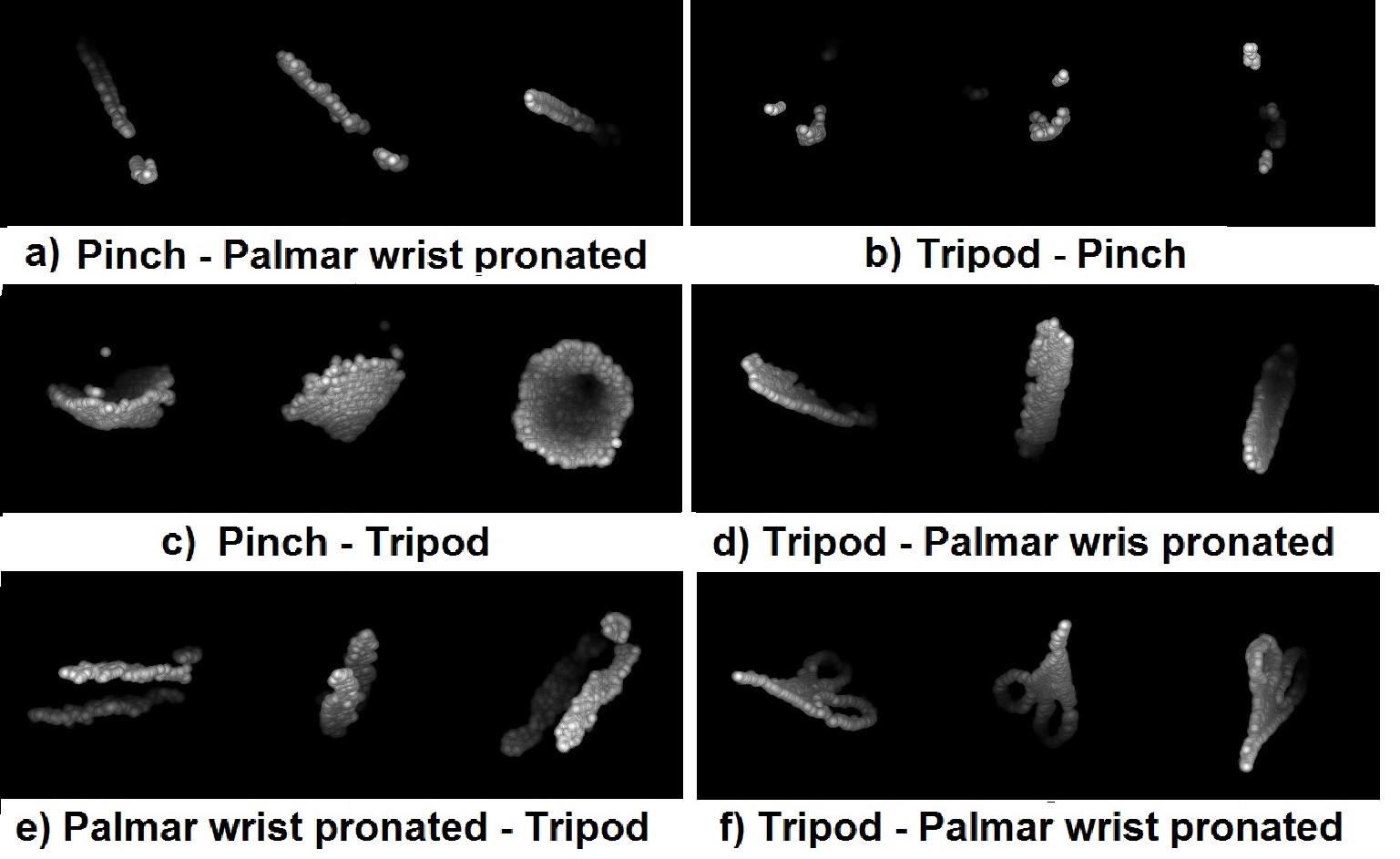}
  \caption{Samples of object point clouds from different views led to incorrect grasp classification. hint: captions are ordered as T-P, where T and P represent true and predicted labels respectively.
  }
  \setlength{\belowcaptionskip}{-45pt}
  \label{figure_6}
  \end{center}
\end{figure}

Some samples of incorrect grasp classification are demonstrated in Figure~\ref{figure_6}. It can be noticed that some errors are happening due to segmentation problems or depth data noise.

\section{Conclusion}
In this paper, an effective and efficient approach for augmenting a hand prosthesis with a depth sensor was presented. Compared to RGB data, depth data provides more shape and grasp relevant information and a depth sensor can be easily mounted on an artificial hand. We added further shape information through estimating surface normals, which led to better grasp estimation performance. Additionally, object segmentation is easier when using depth data. 

There are several barriers in working with the depth data, namely noisy sensor output and extensive computations. The latter can be  eliminated by the use of PointNet, which avoids excessive computations by using point clouds and an efficient architecture. The problem of noise can be overcome by utilizing recent 3-D sensors and noise removal algorithms. Still, depth sensor technologies are improving increasingly and they can be used for commercial artificial hands in near future.

%%%%%%%%%%%%%%%%%%%%%%%%%%%%%%%%%%%%%%%%%%%%%%%%%%%%%%%%%%%%%%%%%%%%%%%%%%%%%%%%

%%%%%%%%%%%%%%%%%%%%%%%%%%%%%%%%%%%%%%%%%%%%%%%%%%%%%%%%%%%%%%%%%%%%%%%%%%%%%%%%

%%%%%%%%%%%%%%%%%%%%%%%%%%%%%%%%%%%%%%%%%%%%%%%%%%%%%%%%%%%%%%%%%%%%%%%%%%%%%%%%
%\section*{APPENDIX}
%\section*{ACKNOWLEDGMENT}
%
%The preferred spelling of the word �acknowledgment� in America is without an �e� after the �g�. Avoid the stilted expression, �One of us (R. B. G.) thanks . . .�  Instead, try �R. B. G. thanks�. Put sponsor acknowledgments in the unnumbered footnote on the first page.

%%%%%%%%%%%%%%%%%%%%%%%%%%%%%%%%%%%%%%%%%%%%%%%%%%%%%%%%%%%%%%%%%%%%%%%%%%%%%%%%

\balance
\bibliographystyle{ieeetr}
\bibliography{Ghazalref}

\begin{thebibliography}{10}

\bibitem{Nazarpour2014}
K.~Nazarpour, C.~Cipriani, D.~Farina, and T.~D. Kuiken, ``Advances in control
  of multi-functional powered upper-limb prostheses,'' {\em IEEE Transactions
  on Neural Systems and Rehabilitation Engineering}, pp.~711--715, 2014.

\bibitem{Saunders2011}
I.~Saunders and S.~Vijayakumar, ``The role of feed-forward and feedback
  processes for closed-loop prosthesis control,'' {\em J Neuroeng Rehabil},
  vol.~8, no.~60, pp.~1--12, 2011.

\bibitem{krasoulis2017improved}
A.~Krasoulis, I.~Kyranou, M.~S. Erden, K.~Nazarpour, and S.~Vijayakumar,
  ``Improved prosthetic hand control with concurrent use of myoelectric and
  inertial measurements,'' {\em Journal of Neuroengineering and
  Rehabilitation}, vol.~14, no.~1, p.~71, 2017.

\bibitem{atzori2016deep}
M.~Atzori, M.~Cognolato, and H.~M{\"u}ller, ``Deep learning with convolutional
  neural networks applied to electromyography data: A resource for the
  classification of movements for prosthetic hands,'' {\em Frontiers in
  Neurorobotics}, vol.~10, 2016.

\bibitem{dovsen2010cognitive}
S.~Do{\v{s}}en, C.~Cipriani, M.~Kosti{\'c}, M.~Controzzi, M.~C. Carrozza, and
  D.~B. Popovi{\'c}, ``Cognitive vision system for control of dexterous
  prosthetic hands: experimental evaluation,'' {\em Journal of Neuroengineering
  and Rehabilitation}, vol.~7, no.~1, p.~42, 2010.

\bibitem{ninu2014closed}
A.~Ninu, S.~Dosen, S.~Muceli, F.~Rattay, H.~Dietl, and D.~Farina, ``Closed-loop
  control of grasping with a myoelectric hand prosthesis: Which are the
  relevant feedback variables for force control?,'' {\em IEEE transactions on
  Neural Systems and Rehabilitation Engineering}, vol.~22, no.~5,
  pp.~1041--1052, 2014.

\bibitem{Markovic2014}
M.~Markovic, S.~Dosen, C.~Cipriani, D.~Popovi{\'c}, and D.~Farina,
  ``Stereovision and augmented reality for closed-loop control of grasping in
  hand prostheses,'' {\em Journal of Neural Engineering}, vol.~11, no.~4,
  p.~046001, 2014.

\bibitem{ghazaei2015exploratory}
G.~Ghazaei, A.~Alameer, P.~Degenaar, G.~Morgan, and K.~Nazarpour, ``An
  exploratory study on the use of convolutional neural networks for object
  grasp classification,'' in {\em Intelligent Signal Processing,2nd IET
  International Conference on}, pp.~1--5, 2015.

\bibitem{ghazaei2017deep}
G.~Ghazaei, A.~Alameer, P.~Degenaar, G.~Morgan, and K.~Nazarpour, ``Deep
  learning-based artificial vision for grasp classification in myoelectric
  hands,'' {\em Journal of Neural Engineering}, vol.~14, no.~3, p.~036025,
  2017.

\bibitem{Giordaniello2017}
F.~Giordaniello, M.~Cognolato, M.~Graziani, A.~Gijsberts, V.~Gregori,
  G.~Saetta, A.-G.~M. Hager, C.~Tiengo, F.~Bassetto, P.~Brugger, B.~Caputo,
  H.~M\"{u}ller, and M.~Atzori, ``Megane {P}ro: myo-electricity, visual and
  gaze tracking integration as a resource for dexterous hand prosthetics,'' in
  {\em IEEE International Conference on Rehabilitation Robotics},
  pp.~1148--1153, 2017.

\bibitem{markovic2015sensor}
M.~Markovic, S.~Dosen, D.~Popovic, B.~Graimann, and D.~Farina, ``Sensor fusion
  and computer vision for context-aware control of a multi degree-of-freedom
  prosthesis,'' {\em Journal of Neural Engineering}, vol.~12, no.~6, p.~066022,
  2015.

\bibitem{LeCun1995}
Y.~LeCun and Y.~Bengio, ``Convolutional networks for images, speech, and time
  series,'' {\em The handbook of brain theory and neural networks}, vol.~3361,
  p.~310, 1995.

\bibitem{qi2016pointnet}
C.~R. Qi, H.~Su, K.~Mo, and L.~J. Guibas, ``Pointnet: Deep learning on point
  sets for 3d classification and segmentation,'' {\em arXiv preprint
  arXiv:1612.00593}, 2016.

\bibitem{lai2011large}
K.~Lai, L.~Bo, X.~Ren, and D.~Fox, ``A large-scale hierarchical multi-view
  rgb-d object dataset,'' in {\em Robotics and Automation (ICRA), 2011 IEEE
  International Conference on}, pp.~1817--1824, IEEE, 2011.

\bibitem{WashingtonDataset}
``{Washington RGB-D dataset}.''
  \url{https://rgbd-dataset.cs.washington.edu/index.html}.
\newblock Accessed: 2017-11-01.

\bibitem{bigbird}
``{BigBIRD ((Big) Berkeley Instance Recognition Dataset}.''
  \url{http://rll.berkeley.edu/bigbird/aliases/f186009c8c/}.
\newblock Accessed: 2017-11-01.

\bibitem{RusuDoctoralDissertation}
R.~B. Rusu, {\em Semantic 3D Object Maps for Everyday Manipulation in Human
  Living Environments}.
\newblock PhD thesis, Computer Science department, Technische Universitaet
  Muenchen, Germany, October 2009.

\bibitem{Krizhevsky2012}
A.~Krizhevsky, I.~Sutskever, and G.~E. Hinton, ``Imagenet classification with
  deep convolutional neural networks,'' in {\em Advances in Neural Information
  Processing Systems}, pp.~1097--1105, 2012.

\bibitem{van2014analysis}
J.~van Doorn, ``Analysis of deep convolutional neural network architectures,''
  2014.

\bibitem{ioffe2015batch}
S.~Ioffe and C.~Szegedy, ``Batch normalization: Accelerating deep network
  training by reducing internal covariate shift,'' in {\em International
  Conference on Machine Learning}, pp.~448--456, 2015.

\end{thebibliography}

\end{document}